# Understanding Sarcoidosis Using Large Language Models and Social Media Data

Nan Miles Xi [1], Hong-Long Ji [2,3], Lin Wang [4*]

[1] Data and Statistical Sciences, AbbVie Inc, North Chicago, IL, 60064

[2] Department of Surgery, Stritch School of Medicine, Loyola University Chicago, Maywood, IL 60153

[3] Burn and Shock Trauma Research Institute, Stritch School of Medicine, Loyola University Chicago, Maywood, IL 60153

[4] Department of Statistics, Purdue University, West Lafayette, IN 47907

* Correspondence: linwang@purdue.edu

## Abstract

Sarcoidosis is a rare inflammatory disease characterized by the formation of granulomas in various organs. The disease presents diagnostic and treatment challenges due to its diverse manifestations and unpredictable nature. In this study, we employed a Large Language Model (LLM) to analyze sarcoidosis-related discussions on the social media platform Reddit. Our findings underscore the efficacy of LLMs in accurately identifying sarcoidosis-related content. We discovered a wide array of symptoms reported by patients, with fatigue, swollen lymph nodes, and shortness of breath as the most prevalent. Prednisone was the most prescribed medication, while infliximab showed the highest effectiveness in improving prognoses. Notably, our analysis revealed disparities in prognosis based on age and gender, with women and younger patients experiencing good and polarized outcomes, respectively. Furthermore, unsupervised clustering identified three distinct patient subgroups (phenotypes) with unique symptom profiles, prognostic outcomes, and demographic distributions. Finally, sentiment analysis revealed a moderate negative impact on patients' mental health post-diagnosis, particularly among women and younger individuals. Our study represents the first application of LLMs to understand sarcoidosis through social media data. It contributes to understanding the disease by providing data-driven insights into its manifestations, treatments, prognoses, and impact on patients' lives. Our findings have direct implications for improving personalized treatment strategies and enhancing the quality of care for individuals living with sarcoidosis.

# Introduction

Sarcoidosis is a rare inflammatory disease characterized by the formation of granulomas in various organs of the body [1,2]. These granulomas are clumps of inflammatory cells that primarily affect the lungs and lymph nodes but can also impact organs such as the eyes, skin, heart, and nervous system [3]. The cause of sarcoidosis is linked to an abnormal immune response to unknown substances in genetically susceptible individuals. According to the American Lung Association, sarcoidosis is a rare disease, with 150,000-200,000 cases reported in the United States per year and approximately 27,000 new cases annually [4]. The occurrence of sarcoidosis varies widely depending on factors such as geography, ethnicity, and environmental exposures. In the United States, sarcoidosis is more common in certain populations, particularly African Americans and people of Northern European descent [5]. Sarcoidosis can manifest with a wide range of symptoms, including persistent cough, shortness of breath, fatigue, weight loss, fever, skin rashes, joint pain, and eye irritation or vision problems [6]. While sarcoidosis is considered a rare disease, it can have a significant impact on affected individuals and their families due to its unpredictable nature and potential for long-term complications. In severe cases, sarcoidosis can lead to organ damage, requiring medical intervention and ongoing monitoring [7]. Symptoms of sarcoidosis can limit a person's ability to work, exercise, or engage in social activities, resulting in chronic damage to patients' mental health.

The rarity of sarcoidosis presents several challenges for patients, healthcare providers, and researchers [8]. Sarcoidosis is often misdiagnosed or diagnosed late, as its symptoms closely resemble those of more common diseases. Healthcare providers may not be familiar with sarcoidosis, leading to delays in diagnosis and appropriate treatment. Due to its infrequency, there is less real-world data on the disease compared to more common conditions. This can result in a lack of understanding of the disease mechanism. The general public may not be familiar with sarcoidosis or may underestimate its severity, leading to misunderstandings about the disease and its impact on patients' lives. The limited research and clinical data available may translate into fewer treatment options for sarcoidosis compared to other diseases. Living with a rare chronic condition like sarcoidosis can significantly affect patients' mental health and overall quality of life. Feelings of isolation, frustration, and anxiety are common among patients, particularly when they encounter difficulties in accessing accurate information, support, and understanding from others [9].

Social media holds the potential to tackle various challenges associated with the rarity of sarcoidosis [10]. These platforms offer a space for individuals affected by the condition to connect with others who understand their experiences, serving as invaluable support networks. Moreover, they serve as rich repositories of information, providing access to the latest insights on sarcoidosis. By leveraging real-world data shared by patients, social media platforms offer valuable insights into the symptoms, treatments, side effects, and prognosis of the disease, thereby contributing to a deeper understanding of its mechanisms and facilitating the development of new therapies. However, effectively collecting and analyzing data from social media poses challenges for the research community [11]. Much of the data on these platforms comprises unstructured text, including posts, comments, and messages. Analyzing this text requires specialized techniques to extract meaningful information, identify sentiment, and categorize content based on its context. Moreover, social media conversations often feature context-specific language, slang, and abbreviations, which can complicate accurate interpretation. Therefore, traditional manual extraction methods struggle to handle the vast amount of information generated at a rapid pace on these platforms.

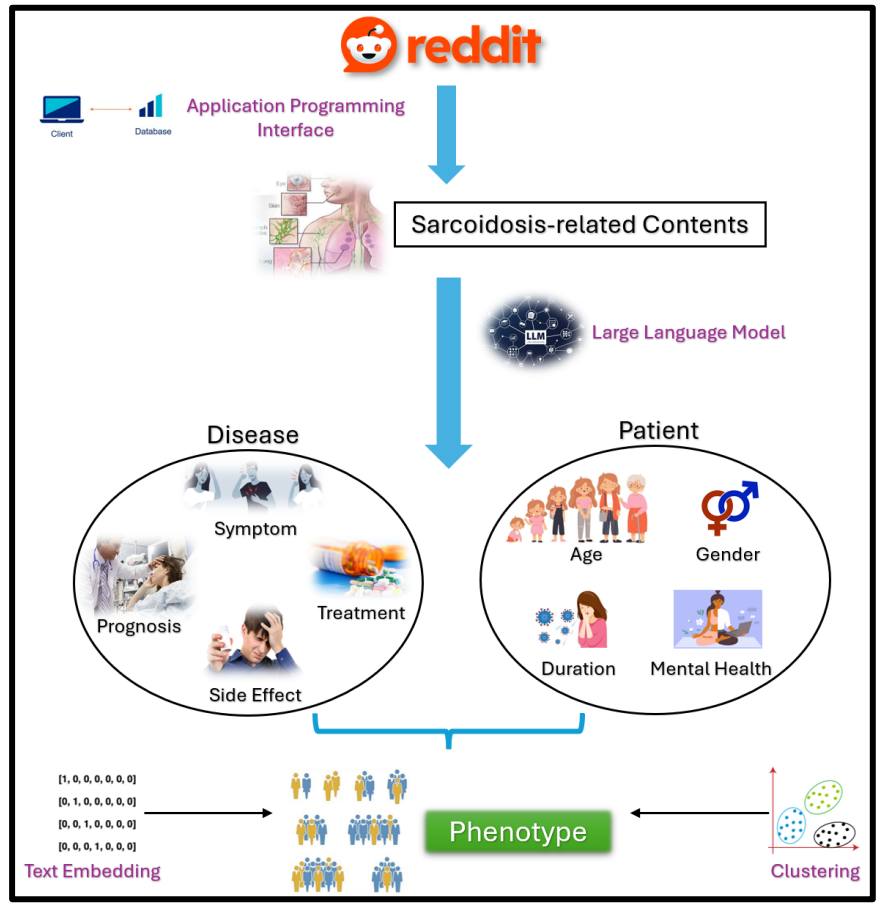

**Figure 1. Schematic representation of the proposed research methodology.**

In this study, we introduce an innovative approach to extracting and analyzing sarcoidosis-related information from the social media platform Reddit, employing Large Language Models (LLMs) (**Figure 1**). LLMs are advanced artificial intelligence models designed to comprehend and generate human-like text [12]. Trained on extensive textual data, these models learn the patterns of natural language, allowing them to convert unstructured text into organized tabular data. We collected textual data from Reddit's sarcoidosis forum and utilized the Generative Pre-trained Transformer (GPT) [13] LLM to understand the diagnosis, symptoms, treatment, side effects, and prognosis of the disease. Additionally, we analyzed demographics, phenotypes, and the mental health status of sarcoidosis patients. In our study, we addressed the following research questions and applied appropriate methods to answer them (**Figure 2**). First, which texts explicitly disclose the author's sarcoidosis diagnosis? To address this, we designed specific prompts for LLMs and obtained binary responses for each downloaded text. Second, what symptoms and treatments are associated with sarcoidosis patients? We utilized LLMs to extract symptoms, treatments, prognosis, and side effects from the texts, guided by predefined lists validated by existing medical literature. Third, what demographic information can be identified for sarcoidosis patients? We crafted prompts to extract demographic details from the texts and standardized the output format for consistency. Fourth, what phenotypes are observed among sarcoidosis patients? We converted patient-related text into numerical embeddings and applied unsupervised clustering to identify distinct patient phenotypes. Finally, what is the impact of sarcoidosis on patients' mental health? We used LLMs to classify text sentiment into three categories and compared sentiment shifts before and after the reported sarcoidosis diagnosis.

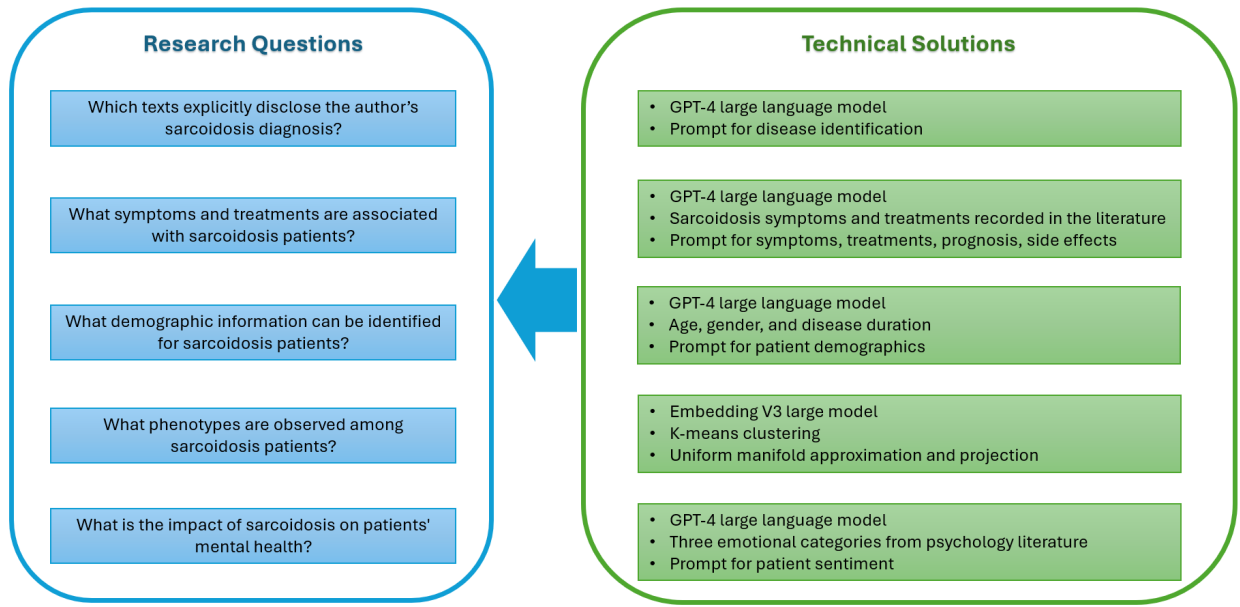

**Figure 2. Research questions addressed in this study and their corresponding technical solutions.**

Our results demonstrate the LLM's high accuracy in identifying sarcoidosis diagnoses and corresponding symptoms. We find prednisone emerged as the most prescribed medication, followed by methotrexate and infliximab. Prognosis analysis revealed a predominantly positive outlook, while patients reported mild yet notable treatment-related side effects such as weight gain, insomnia, and diabetes. Notably, younger patients exhibited higher rates of both good and poor prognoses, with females generally experiencing better outcomes compared to males. Patients with a good prognosis tended to have shorter disease durations than those with moderate or poor prognoses. Our analysis identified three distinct patient subgroups (phenotypes) characterized by unique symptom profiles, prognostic outcomes, and demographic distributions. Finally, we observed a moderate impact of sarcoidosis on patients' mental well-being, evidenced by shifts in both negative and positive content following diagnosis, with a more pronounced emotional effect observed among female and younger patients.

Our work has made several contributions to health informatics research. First, by leveraging social media data and advanced language models, the research provides data-driven insights into sarcoidosis, enriching our understanding of the disease's manifestations and management. Second, the findings have direct implications for clinical practice, which can improve personalized treatment strategies, patient outcomes, and the quality of care for individuals living with sarcoidosis. Third, we have collected real-world evidence about the treatment effects and patient experience. This real-world evidence complements traditional clinical research data, offering a more holistic understanding of the disease and its impact on patients' lives. Fourth, the study highlights disparities in disease prognosis based on demographic characteristics. This understanding is crucial for developing more equitable healthcare practices that account for the diverse needs of patients. Fifth, the identified sarcoidosis phenotypes offer a new perspective on the heterogeneity of sarcoidosis, showing the potential for personalized medicine approaches to improve patient care. To our knowledge, this is the first study to categorize sarcoidosis patients into phenotypes using LLMs and social media data. Last, our sentiment analysis reveals the mental health impact of sarcoidosis, particularly among female and younger patients. It underscores the need for psychological support as part of comprehensive disease management. Overall, our work contributes to the growing body of literature on social media's role in health communication and patient support, highlighting its potential as a rich data source for understanding disease experiences, improving healthcare delivery, and fostering community engagement and support.

## Related work

The research community has shown a growing interest in leveraging social media platforms for studying rare diseases. Coulson et al. examined the dynamics of social support within an online support group for Huntington's disease [14]. Subirats et al. conducted a content-based and temporal analysis on Facebook, aiming to enhance engagement among rare disease patients and assist rare disease organizations in aligning their priorities with the interests expressed in social networks [15]. Akre et al. performed a pilot study to evaluate a dedicated social media website catering to parents of adolescents with neurofibromatosis type 1 [16]. Ali et al. utilized social media to collect data from patients with Dravet syndrome, investigating post-surgical seizure reduction and overall satisfaction with varus nerve stimulation [17]. Applequist et al. developed a tailored social media platform to recruit participants for a clinical trial from rare disease populations [18]. Bi et al. explored topic categorizations and sentiment polarity for albinism in an online health forum [19]. Canalichio et al. employed social media to survey adult females with exstrophy, aiming to assess long-term patient-reported sexual, reproductive, and continence outcomes. [20] Shen et al. investigated the clinical effects of WeChat-based peri-operative care on parents of children with congenital megacolon [21]. Pemmaraju et al. conducted an analysis based on Twitter to understand trends and topics within the community of myeloproliferative neoplasms [22].

In parallel, utilizing Large Language Models (LLMs) for medical data analysis has gained significant attention. Guo et al. utilized LLMs to detect Covid-19 cases and extract disease symptoms from online forums [23]. Singhal et al. conducted a benchmark study demonstrating LLMs' substantial medical knowledge and reasoning capabilities [24]. Gilson et al. showed that LLMs achieve results equivalent to third-year medical students in medical knowledge exams [25]. Ayers et al. conducted a comparative study revealing a preference for LLM-generated responses over those from licensed physicians when addressing patient queries [26]. Yang et al. developed an LLM for various medical tasks, including clinical concept extraction and medical question answering [27]. Krusche et al. applied LLMs in diagnosing inflammatory rheumatic diseases, demonstrating improved sensitivity compared to rheumatologists [28]. Van et al. employed LLMs for text simplification, aiding in translating complex medical reports into understandable language for patients [29]. Fitzpatrick et al. used LLM-based conversational agents to deliver cognitive behavior therapy to young adults with depression and anxiety symptoms [30]. Shyr et al. systematically explored the design of model prompts in identifying disease phenotypes with LLMs [31].

Our work novelly integrates cutting-edge LLM methodology and a moderate-sized volume of patient self-reported data on social media. To our best knowledge, this study represents the first application of LLMs to understand sarcoidosis through social media data. Consequently, our research fills a critical gap in the existing literature, opening up promising avenues for future investigations.

## Methods

**Data collection.** The dataset analyzed in this study comprises threads and comments from the sarcoidosis forum on the social media platform Reddit (www.reddit.com). A thread refers to the initial post made by a user and typically represents a topic of discussion or a piece of content that other users can interact with. A comment refers to a response made by a user to a thread or to another comment within the thread. We accessed a user-generated and topic-specific forum known as r/sarcoidosis. We retrieved the relevant data using the R package RedditExtractoR [32], which interfaces with the Reddit application programming interface (API) [33]. The dataset contains threads and comments spanning from the inception of the sarcoidosis forum in September 2012 to March 2024. The information in our dataset is publicly available on Reddit and is generally anonymized. Therefore, no identifying information about the authors is included. Additionally, no instances were found where authors provided Protected Health Information (PHI) that could lead to personal identification. Throughout this process, we have followed

Reddit's terms, user agreement, and conditions regarding data mining and user privacy. In total, we downloaded 801 threads and 8197 comments.

**Disease and patient annotation.** The threads and comments in our dataset are not necessarily related to sarcoidosis. We need to examine if the author disclosed a diagnosis of sarcoidosis in the text they posted. To achieve this, we employed OpenAI's GPT-4 Turbo model [13] to automatically annotate each thread and comment. A specific prompt was formulated to address this task: 'Does the author of the following text explicitly indicate that himself/herself has been diagnosed with sarcoidosis? Answer yes, no, or unclear'. This prompt was integrated with the text of each thread and comment, and the resulting model responses were recorded. The annotations were carried out utilizing OpenAI's API with the model version gpt-4-0125-preview. To maintain consistency and reproducibility, we set the model temperature to 0, ensuring that GPT-4 selected the most probable token in its responses. The subsequent analysis was conducted using the same model version and parameter settings. Threads and comments that received a model answer of "yes" were identified as text relating to sarcoidosis and were grouped based on their authors, whom we subsequently designated as sarcoidosis patients.

**Symptom and treatment identification.** For each sarcoidosis patient, we employed the GPT-4 Turbo model to identify disease symptoms in the text they authored. We compiled a list of 28 sarcoidosis symptoms documented in medical literature [6]. These symptoms were classified into six categories: general symptoms, pulmonary symptoms, ocular symptoms, dermatological symptoms, cardiac symptoms, and neurological symptoms (**Supplementary Table S1**). We designed the following prompt to inquire about the model for disease symptoms: 'Which of the listed symptoms are described in the following text? If no symptoms are mentioned, output "no symptom". The symptoms are … The text is …'. Here, our analysis focused solely on text annotated as relevant to sarcoidosis for each patient. We also investigated the treatments utilized by patients to manage the disease, gathering information on 24 sarcoidosis medications based on clinical practice [34] (**Supplementary Table S2**). To extract treatment details from the text, we crafted the following prompt: 'Which of the listed treatments are used in the following text? If no treatments are used, output "no treatment". The treatments are … The text is …'. Again, the text considered was exclusively related to sarcoidosis for each patient.

**Prognosis and side effect analysis.** We categorized the prognosis of sarcoidosis into three levels: good, moderate, and poor. To determine the prognosis level of each patient, we presented the following prompt to the model: 'The author of the following text was diagnosed with sarcoidosis. Summarize the prognosis, choosing from good, moderate, or poor. If there is no prognosis information, output unknown. The text is …'. Furthermore, we examined the side effects experienced during sarcoidosis treatments for each patient. We compiled a list of 17 side effects documented in the medical literature [35] (**Supplementary Table S2**) and identified which ones were mentioned in the patients' threads and comments by asking: 'Which side effects during sarcoidosis treatments are described in the following text? Output the exact side effects from the list … If there is no side effect, output "no side effect". The text is …'. Similar to the previous section, only the threads and comments related to sarcoidosis were considered for each patient.

**Demographics extraction.** We extracted three demographic pieces of information – age, gender, and duration of sarcoidosis – for each patient from their posted threads and comments. Our prompt to the model was: 'The author of the following text was diagnosed with sarcoidosis. Summarize his/her age, gender, and duration since diagnosis. If there is no information for one item, output unknown for that item. The text is …'. To standardize the model output of disease duration, we converted it to the unit of years. For instance, if the model identified the duration as three months, we transformed it to 0.25 years. Importantly, the demographic information obtained from the patient's text is very basic, thus making it impossible to identify any personal privacy from this data.

**Patient clustering analysis.** We employed unsupervised clustering to identify patient subgroups based on their disease and demographic characteristics. Initially, we selected patients with complete information on symptoms, age, gender, and disease duration. Subsequently, we concatenated this information to create a string of patient records. Next, we utilized OpenAI's Embedding V3 large model [36] to transform these textual patient records into a 3072-dimensional numerical embedding vector. This embedding representation measures the semantic similarity between different patient records, whereby patients with similar disease and demographic characteristics are closer in this embedding space. To reduce the dimensionality of the embedding while preserving semantic similarity and avoiding the "curse of dimensionality," we employed Uniform Manifold Approximation and Projection (UMAP) [37] to reduce the dimensionality to five. Finally, we performed k-means clustering on the five-dimensional embeddings of patient records to identify patient subgroups. The k-means clustering was implemented using the R function kmeans, with the parameter nstart set to 25. The number of clusters (k) was selected based on the majority vote of 30 indices calculated by the function NbClust in the R package NbClust [38].

**Sentiment analysis.** We collected all the threads authored by sarcoidosis patients on Reddit and conducted sentiment analysis. We summarized three emotional categories from psychology literature: positive, negative, and neutral [39]. To determine the sentiment of each thread, we designed the following prompt for the GPT-4 Turbo model: 'What is the sentiment of the following text? Choose from positive, negative, or neutral. The text is …'. For each patient, we identified the timestamp when they first mentioned the diagnosis of sarcoidosis. While some patients indicated a long history of diagnosis in their first posts, most shared their text very close to their diagnosis. Therefore, these timestamps served as proxies for indicating each patient's diagnosis time. Utilizing these diagnosis times, we divided each patient's thread into two segments: one preceding the diagnosis and the other subsequent to it.

**Mislabeling and sensitivity analysis.** Similar to other machine learning models, LLMs can introduce mislabeling issues. The impact of this issue can be measured by performing manual validation and constructing confidence intervals for findings in our study. First, we conducted a manual validation to estimate the model's performance in identifying sarcoidosis diagnosis (**Supplementary File validation.csv**). Second, we used the results from manual validation to estimate the model's overall accuracy, sensitivity, specificity, true positive value (TPV), and true negative value (TNV). Third, we calculate the 95% confidence intervals (CIs) for the TPV and TNV using the following formula:

$$CI_{TPV} = \widehat{TPV} \pm Z \times \sqrt{\frac{\widehat{TPV}(1-\widehat{TPV})}{n}}$$

$$CI_{TNV} = \widehat{TNV} \pm Z \times \sqrt{\frac{\widehat{TNV}(1-\widehat{TNV})}{n}}$$

where $\widehat{TPV}$ and $\widehat{TNV}$ are the estimations from manual validation, $n$ is the sample size, and Z is the Z-score for the 95% interval. Next, based on the definitions of TPV, TNV, and false negative value (FNV):

$$TPV = \frac{True\ Positve\ (TP)}{Predicted\ Positive\ (PP)}$$

$$TNV = \frac{True\ Negative\ (TN)}{Predicted\ Negative\ (PN)}$$

$$FNV = \frac{False\ Neagtive\ (FN)}{Predicted\ Negative\ (PN)} = 1 - TNV$$

the 95% CIs for the TP and FN can be calculated by

$$CI_{TP} = CI_{TPV} \times PP$$

$$CI_{FN} = CI_{FNV} \times PN = (1 - CI_{TNV}) \times PN$$

Finally, for the model annotation T

$$T = TP + FN$$

its 95% CI can be calculated by combining the lower and upper bounds of $CI_{TP}$ and $CI_{FN}$:

$$CI_T = (l_{CI_{TP}} + l_{CI_{FN}},\ u_{CI_{TP}} + u_{CI_{FN}})$$

## Results

**The large language model demonstrated high accuracy in identifying content indicating sarcoidosis diagnosis.** Among the 801 threads and 8197 comments posted in the r/sarcoidosis forum, we identified 371 threads and 2106 comments where authors disclosed their sarcoidosis diagnosis. These contents originated from 679 unique authors, whom we categorized as sarcoidosis patients for our analysis (column "patients" in **Supplementary File sarc.patients.csv**). To validate the model's accuracy in identifying sarcoidosis diagnosis, we randomly sampled 200 threads and 100 comments, and one author manually annotated the textual contents. Comparing these annotations to the human-annotated ground truth, we found the model achieved an overall accuracy of 0.9394. Additionally, sensitivity, specificity, true positive value, and true negative value were calculated as 0.9329, 0.9474, 0.9563, and 0.9197, respectively (**Supplementary File validation.csv**). This validation underscores the model's ability to accurately identify text indicating sarcoidosis diagnosis. **Table 1** shows textual examples from our dataset along with corresponding model and human annotations. Given the rarity of sarcoidosis, our findings represent a relatively moderate-sized patient cohort.

**Table 1. Textual examples from Reddit text, along with corresponding model and human annotations for sarcoidosis diagnosis.**

| | Text | Model annotation | Human annotation |
|---|---|---|---|
| **Correct (93.94%)** | Hi, recently I have been diagnosed with pulmonary sarcoidosis. My symptoms are mild right now. like coughing, tiredness, and very rarely shortness of breath… | Yes | Yes |
| | I'm wanting to do everything I can to get my sarcoid into remission without too many side effects from prednisone. I have a stressful full-time desk job with a large company in the US… | Yes | Yes |
| | Just saw this on X and might be interesting for some people in this forum. | No | No |
| | I got the flu Nov 2022. It affected my lower back so badly. I couldn't walk for several weeks. I've since had radio frequency ablations done… | No | No |
| **Wrong (6.06%)** | Hey there guys I'm just wondering if hotspots like right now of typing this one of my ears is red and warm and I'm wondering if that is connected to the sarcoidosis or just random or something Else? | Yes | No |
| | I'm in Kelowna, and was very lucky to get diagnosed right away and be receiving excellent care from my doctor and respirologist. Just wondering how it's going in other parts of the province. | No | Yes |

**Sarcoidosis patients exhibited a wide range of symptoms.** Through an analysis of threads and comments authored by individuals with sarcoidosis, we identified 485 patients who reported 27 different symptoms during the course of their illness (**Figure 3A;** column "symptom" in **Supplementary File sarc.patients.csv**). Overall, general symptoms and pulmonary symptoms are the two most prevalent symptom categories. Among the individual symptoms, the top five most frequently reported are fatigue (35.26%), swollen lymph nodes (31.34%), shortness of breath (24.33%), joint pain (22.27%), and cough (20.00%). These findings align closely with symptoms documented in medical literature [6]: Fatigue emerges as one of the most prevalent symptoms, accounting for approximately 50% of cases; The enlargement of abdominal lymph nodes, present in about 30% of patients, mirrors the 31.34% prevalence of swollen lymph nodes in our dataset; Both shortness of breath and cough, classified as respiratory symptoms, are observed in 30%–53% of patients at presentation. While joint pain is not typically listed among the top-known symptoms, it ranks fourth in our dataset. Conversely, the five least frequent symptoms include fever (3.92%), heart failure (3.92%), red tender bumps on shins (3.92%), seizures (2.68%), and weak or paralyzed facial muscles (2.68%). Fever is considered a non-specific constitutional symptom in sarcoidosis. It may be underreported due to patients attributing it to other common causes. Heart failure is a less common symptom in the literature associated with cardiac sarcoidosis [40]. Weak or paralyzed facial muscles, indicative of Löfgren's syndrome (LS), account for less than 1% of cases [6]. Red tender bumps on shins and seizures are also infrequently documented in medical literature. Our findings offer valuable firsthand data on sarcoidosis symptoms, facilitating prompt recognition of this rare disease among the general public.

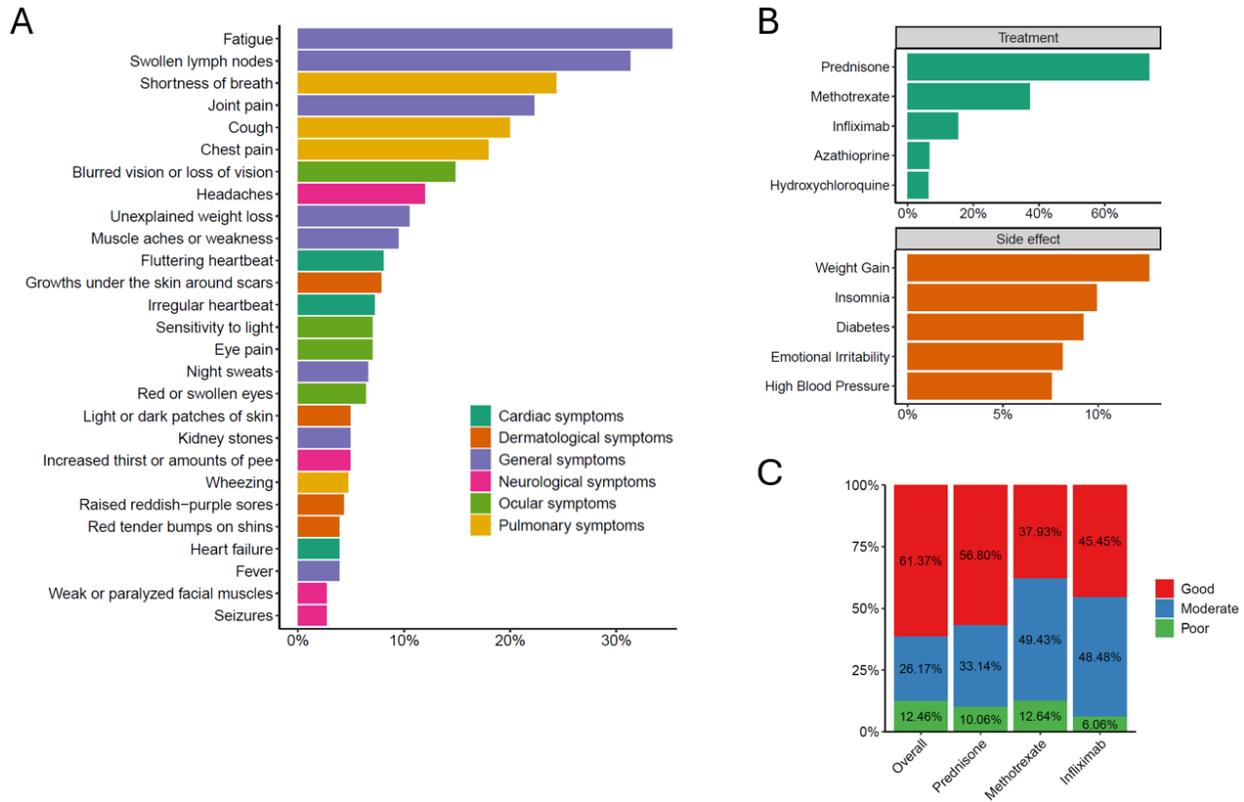

**Figure 3. Symptom, treatment, side effects, and prognosis of sarcoidosis patients. A.** 27 symptoms reported during the course of illness (n=485). **B.** Top 5 most prescribed medications (n=365) and top 5 most frequent side effects (n=201) during treatment. **C.** Prognosis of all patients (n=321) and those who used top 3 medications.

**Treatment, prognosis, and side effects.** Sarcoidosis treatment primarily focuses on symptom management and the prevention of organ damage. Medications used for this purpose target inflammation by modulating or suppressing the patient's immune response. Our analysis of patient-reported data revealed the utilization of 21 medications for sarcoidosis treatment (**Table 2;** column "treatment" in **Supplementary File sarc.patients.csv**). Prednisone emerged as the most prescribed medication (73.42%), followed by methotrexate (37.26%), infliximab (15.34%), azathioprine (6.58%), and hydroxychloroquine (6.30%) (**Figure 3B**). Medications such as prednisone and methotrexate are frequently used as first-line therapies to reduce inflammation and manage associated symptoms [41]. In contrast, treatments like infliximab are often reserved for second- or third-line therapy, particularly in cases of refractory sarcoidosis [42]. Additionally, we examined the prognosis of sarcoidosis treatment using a three-tiered system—good, moderate, and poor. Textual analysis successfully identified prognosis information for 321 patients, revealing that 61.37% claimed a good prognosis, 26.17% reported a moderate prognosis, and 12.46% indicated a poor prognosis (**Figure 3C;** column "prognosis" in **Supplementary File sarc.patients.csv**). **Supplementary Table S3** provides examples of text annotated by the model across these prognosis levels. Furthermore, we investigated treatment-related side effects reported by patients in their Reddit posts (**Table 2**). Among 201 patients who reported side effects, the top five most frequent were weight gain (45.77%), insomnia (35.82%), diabetes (33.33%), emotional irritability (29.35%), and high blood pressure (27.36%) (**Figure 3B**). Although generally mild, some less frequent side effects, such as cataracts (10.45%) and glaucoma (8.96%), may lead to more severe consequences.

**Table 2. Treatment and related side effects reported by sarcoidosis patients.**

| Treatment | Percentage (%) n=365 | Count | Side effect | Percentage (%) n=201 | Count |
|---|---|---|---|---|---|
| Prednisone | 73.42 | 268 | Weight gain | 45.77 | 92 |
| Methotrexate | 37.26 | 136 | Insomnia | 35.82 | 72 |
| Infliximab | 15.34 | 56 | Diabetes | 33.33 | 67 |
| Azathioprine | 6.58 | 24 | Emotional irritability | 29.35 | 59 |
| Hydroxychloroquine | 6.30 | 23 | High blood pressure | 27.36 | 55 |
| Implanted cardiac pacemaker | 5.48 | 20 | Osteoporosis | 26.37 | 53 |
| Steroids | 5.21 | 19 | Depression | 22.89 | 46 |
| Prednisolone | 4.93 | 18 | Nausea | 22.39 | 45 |
| Humira | 4.11 | 15 | Skin Bruising | 21.39 | 43 |
| Cortisone | 3.29 | 12 | Dizziness | 16.42 | 33 |
| Adalimumab | 2.19 | 8 | Thrush | 16.42 | 33 |
| Nonsteroidal anti-inflammatory drugs | 2.19 | 8 | Gastrointestinal tract problems | 12.44 | 25 |
| Leflunomide | 1.92 | 7 | Diarrhea | 10.95 | 22 |
| Physical therapy | 1.92 | 7 | Acne | 10.45 | 21 |
| Naproxen | 1.10 | 4 | Cataracts | 10.45 | 21 |
| Rituximab | 1.10 | 4 | Hoarse voice | 10.45 | 21 |
| Methylprednisolone | 0.82 | 3 | Glaucoma | 8.96 | 18 |
| Mycophenolate | 0.82 | 3 | | | |
| Corticotropin | 0.55 | 2 | | | |
| Hydrocortisone | 0.55 | 2 | | | |
| Plaquenil | 0.55 | 2 | | | |

We also examined the treatment effects of the three most used sarcoidosis medications—prednisone, methotrexate, and infliximab (**Figure 3C;** column "side_effect" in **Supplementary File sarc.patients.csv**). Among patients using these medications, infliximab demonstrated the highest percentage of good or moderate prognosis (94.11%), followed by prednisone (89.94%) and methotrexate (87.50%). Prednisone exhibited the highest percentage of good prognosis (56.80%), followed by infliximab (44.11%) and methotrexate (37.50%). Overall, these medications displayed strong effectiveness in suppressing sarcoidosis symptoms. Moreover, we examined the most frequent side effects associated with these medications (**Table 3 and Figure 3B**). Weight gain, insomnia, and diabetes were commonly reported across all three medications, while emotional irritability and depression were unique to patients using prednisone and infliximab, respectively.

Table 3. Side effects of the top 3 most prescribed medications.

| Medication | Side effect | Percentage (%) | Count |
|---|---|---|---|
| Prednisone (n=147) | Weight gain | 54.42 | 80 |
| | Insomnia | 41.50 | 61 |
| | Diabetes | 39.46 | 58 |
| | Emotional irritability | 35.37 | 52 |
| | High blood pressure | 32.65 | 48 |
| Methotrexate (n=79) | Weight gain | 53.16 | 42 |
| | Nausea | 40.51 | 32 |
| | Diabetes | 35.44 | 28 |
| | Insomnia | 35.44 | 28 |
| | High blood pressure | 34.18 | 27 |
| Infliximab (n=21) | Weight gain | 57.14 | 12 |
| | Nausea | 47.62 | 10 |
| | Depression | 42.86 | 9 |
| | Diabetes | 42.86 | 9 |
| | Insomnia | 42.86 | 9 |

**Relationship between patient demographics and disease prognosis.** The association between demographic characteristics of sarcoidosis patients and disease prognosis has been well-documented in medical literature [43,44]. Leveraging social media data to explore this relationship provides real-world evidence to enhance treatment outcomes for patients. In our study, we collected demographic data, including age, gender, and disease duration, from patient text records (columns "age", "gender", and "duration" in **Supplementary File sarc.patients.csv**). Among 121 patients with gender information, 60 are males and 61 are females. Our analysis revealed that among younger patients (age ≤ 35), both good and poor prognosis rates were higher, at 58.53% and 19.51%, respectively, compared to their older counterparts (age > 35; good=42.30%, poor=13.46%) (**Figure 4A**). We selected 35 as the age threshold because the median ages in our dataset and the population-based study by Ungprasert et al. [5] are 36 and 37.7 years, respectively (**Supplementary Figure S2**). We then rounded them to 35 for simplicity. This threshold was determined based on the distribution of our samples and does not represent a general definition of "young" or "old." This result indicates that prognosis tends to be more polarized in younger patients, which was not well documented in prior sarcoidosis research. This insight suggests age-specific variability in disease progression, which can inform tailored treatment strategies.

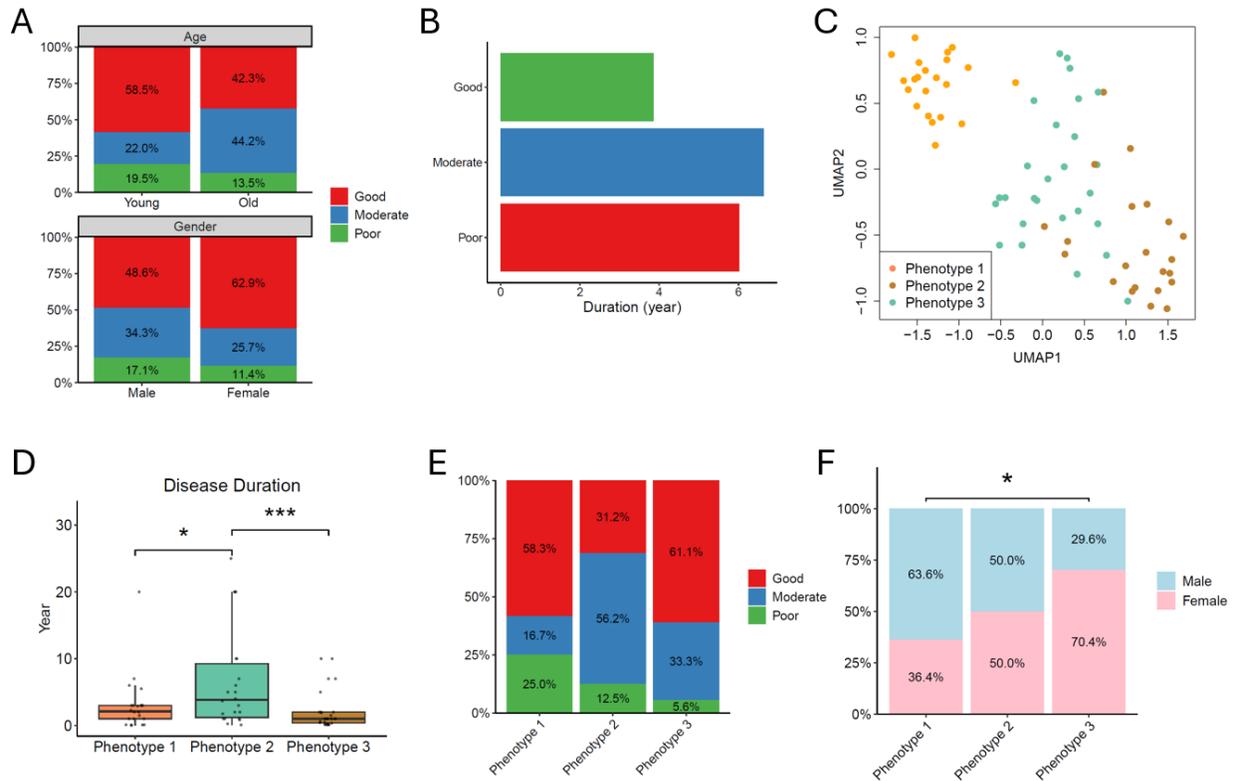

**Figure 4. Patient demographics, prognosis, disease duration, and phenotype. A.** Prognosis distributions of different age and gender groups. **B.** Average disease duration in each prognosis category. **C**. UMAP plot of three patient phenotypes. **D**. Disease duration of three phenotypes. * p-value < 0.1, ** p-value < 0.05, *** p-value < 0.01, Wilcoxon rank-sum test. **E**. Prognosis distribution of three phenotypes. **F**. Gender distribution of three phenotypes. * p-value < 0.1, $\chi^2$ test.

Furthermore, we observed a significant difference in prognosis between genders. Female patients demonstrated a notably better prognosis, with 62.85% exhibiting good outcomes compared to 48.57% in male patients. Conversely, the percentage of poor prognosis was lower among females (11.42%) compared to males (17.13%) (**Figure 4A**). Additionally, we found a correlation between prognosis and disease duration. Patients with a good prognosis had an average disease duration of 3.86 years, which was significantly shorter than the averages of 6.64 years and 6.02 years among the moderate and poor prognosis groups, respectively (**Figure 4B**). This underscores the importance of early diagnosis and prompt initiation of treatment, especially considering that chronic sarcoidosis tends to be more challenging to manage [45].

**Relationship between symptom patterns and treatment outcomes.** We identified patients with both symptom and prognosis information (n=269) and analyzed their relationships from three perspectives. First, we analyzed the relationship between the number of symptoms reported by patients and their prognosis. Our findings indicate a negative association, where an increase in the number of symptoms correlates with a poorer prognosis. This relationship was statistically significant (**Supplementary Figure S1A and S1B; Supplementary Table S5**). Second, we investigated the impact of the number of symptom categories (**Supplementary Table S1**) on prognosis. Our results show that patients exhibiting symptoms across a larger number of categories tend to have worse prognoses compared to those with symptoms in fewer categories. This difference was also statistically significant (**Supplementary Figure S1C and S1D; Supplementary Table S6**). Third, among patients reporting unique types of symptoms (n=108), we found no significant difference in prognosis across six symptom categories (**Supplementary Table S7**). This suggests that the specific type of symptom may not be as influential on prognosis as the number and variety of symptoms. These results suggest a complex interaction between symptomatology and treatment efficacy, a topic not extensively covered in the existing literature. Our findings offer new insights that could potentially advance the understanding of sarcoidosis mechanisms and inform treatment strategies.

**Phenotypes identification in sarcoidosis patients.** Phenotypes refer to distinct patient subgroups characterized by observable traits associated with a particular disease [46]. These traits contain a wide range of features, including clinical symptoms, disease duration, and patient demographics. Phenotypes provide a means of categorizing sarcoidosis patients based on their clinical manifestations and underlying biological mechanisms. To identify sarcoidosis phenotypes, we concatenated patients' symptoms, age, gender, and disease duration into a textual description, and transformed it into numerical embeddings (Methods). Using unsupervised clustering, we found three patient subgroups (phenotypes) based on their numerical embeddings of textual descriptions (**Figure 4C;** column "phenotype" in **Supplementary File sarc.patients.csv**). We investigated the characteristics of these phenotypes through comparing their corresponding clinical and demographic variables. Patients in phenotype two showed significantly longer disease durations and less good prognose compared to other groups, consistent with our previous findings regarding the inverse relationship between prognosis and disease duration (**Figures 4D and 4E**). Moreover, phenotype three had a notably higher proportion of female patients. Although three phenotypes shared certain common symptoms, such as cough, chest/joint pain, and fatigue, each phenotype also exhibited unique symptoms (**Table 4**). For instance, among the top five most frequent symptoms, chest pain (27.27%) and heart failure (18.18%) were characteristics of phenotype one, while eye problems were predominant in phenotype two (50.00% and 45.45%), and swollen lymph nodes were prevalent in phenotype three (59.26%). Similarly, each phenotype showed distinct side effects from treatment. For example, phenotype one exhibited depression (18.18%), acne (13.64%), and cataracts (13.64%), phenotype two displayed nausea (36.36%), and phenotype three showed incidence of emotional irritability (22.22%) and high blood pressure (18.52%) (**Table 5**). Overall, the identification of phenotypes revealed unique clinical symptoms, prognostic outcomes, and demographic distributions, offering valuable insights into the underlying biological mechanisms of sarcoidosis.

**Table 4. Top 5 most frequent symptoms of three phenotypes.** Symptoms unique to each phenotype are underscored.

| Phenotype 1 (n=22) | | | Phenotype 2 (n=22) | | | Phenotype 3 (n=27) | | |
|---|---|---|---|---|---|---|---|---|
| Symptom | Percentage (%) | Count | Symptom | Percentage (%) | Count | Symptom | Percentage (%) | Count |
| Shortness of breath | 54.55 | 12 | Blurred vision or loss of vision | 50.00 | 11 | Swollen lymph nodes | 59.26 | 16 |
| Cough | 40.91 | 9 | Joint pain | 50.00 | 11 | Fatigue | 51.85 | 14 |
| Chest pain | 27.27 | 6 | Eye pain | 45.45 | 10 | Joint pain | 37.04 | 10 |
| Heart failure | 18.18 | 4 | Fatigue | 45.45 | 10 | Shortness of breath | 29.63 | 8 |
| Fatigue | 13.64 | 3 | Fluttering heartbeat | 45.45 | 10 | Cough | 18.52 | 5 |

**Table 5. Top 5 most frequent side effects of three phenotypes.** Side effects unique to each phenotype are underscored.

| Phenotype 1 (n=22) | | | Phenotype 2 (n=22) | | | Phenotype 3 (n=27) | | |
|---|---|---|---|---|---|---|---|---|
| Side effect | Percentage (%) | Count | Side effect | Percentage (%) | Count | Side effect | Percentage (%) | Count |
| Weight gain | 36.36 | 8 | Weight gain | 40.91 | 9 | Weight gain | 33.33 | 9 |
| Depression | 18.18 | 4 | Insomnia | 36.36 | 8 | Diabetes | 29.63 | 8 |
| Osteoporosis | 18.18 | 4 | Nausea | 36.36 | 8 | Emotional irritability | 22.22 | 6 |
| Acne | 13.64 | 3 | Osteoporosis | 31.82 | 7 | Insomnia | 22.22 | 6 |
| Cataracts | 13.64 | 3 | Diabetes | 27.27 | 6 | High blood pressure | 18.52 | 5 |

**Sarcoidosis showed a moderate impact on patients' mental health conditions.** The symptoms and treatment side effects of sarcoidosis have been observed to negatively affect patients' mental health [47]. To explore these potential impacts using social media data, we conducted an analysis of the sentiment expressed in threads posted by sarcoidosis patients on Reddit, categorizing them as positive, negative, or neutral (**Supplementary Table S4;** column "sentiment" in **Supplementary File thread.all.csv**). These sentiments served as proxies for evaluating the mental well-being of sarcoidosis patients. Overall, the majority of text posted by patients exhibited non-negative sentiment, with 28.69% being positive, 38.52% neutral, and 32.79% negative (**Figure 5A**). Interestingly, we observed a decrease in both negative and positive content following diagnosis (**Figure 5B**), suggesting that patients expressed fewer extreme attitudes in their online posts post-diagnosis. Among patients who posted at least one positive thread and ten total threads before and after diagnosis, the majority (53.33%) posted fewer positive threads post-diagnosis compared to pre-diagnosis. On average, there was a 13.45% decrease in positive threads among patients post-diagnosis, significantly higher than the average increase of 9.94% (**Figure 5C**). Similarly, among patients who posted at least one negative thread and ten total threads before and after diagnosis, the majority (52.25%) posted more negative threads post-diagnosis compared to pre-diagnosis. The average increase in negative threads among patients post-diagnosis was 11.83%, slightly higher than the average decrease of 11.78% (**Figure 5C**). These findings suggest a moderate negative impact on patients' mental health as reflected by the sentiment in their online posts.

Furthermore, we observed that the impact of diagnosis was associated with patients' gender and age. Female patients experienced an average decrease of 10.84% in positive content post-diagnosis, while male patients exhibited a slight increase of 0.44%. Female patients also showed a larger increase in negative content post-diagnosis compared to males (2.72% vs. 0.06%) (**Figure 5D**). Similarly, patients younger than 35 years old experienced an average decrease of 3.96% in positive content post-diagnosis, while older patients saw an increase of 3.48%. In terms of negative content, younger patients exhibited an average increase of 1.33%, higher than the 0.04% increase observed in older patients (**Figure 5E**). Overall, sentiment analysis indicated that female and younger patients tended to be more impacted than male and older patients, as evidenced by the content they posted on Reddit.

## Discussion and Limitations

**Impact of model mislabeling.** The results of our study depend on the LLM's ability to accurately extract relevant content from text data. Mislabeling by LLMs can potentially introduce bias into the analysis. To assess this impact, we conducted a sensitivity analysis by constructing confidence intervals (CIs) for the results. **Supplementary Table S14** presents the 95% CIs for the top three treatments and side effects, following the methods proposed in the Methods section. The widths of CIs range from 10-20 for absolute counts and 5-10 for percentages, which indicates a relatively accurate model estimation. We anticipate narrower CIs, and thus more precise results, with larger sample sizes and more advanced LLMs (e.g., GPT-4). Our proposed method is applicable to most results in our study, providing a measure of the impact of mislabeling. Additionally, addressing the sensitivity analysis for unsupervised clustering (patient phenotype) remains challenging. One potential approach is to construct CIs for the variables used in k-means clustering and then repeat clustering with different combinations of variable values to test the stability of the clustering results. The more stable the clustering, the less impact mislabeling will have on phenotype identification. Due to space limitations, we plan to explore these analyses in a future study.

The text analyzed in this study should be authored directly by patients, not by their family members or friends. To minimize the inclusion of third-party narratives, we designed the prompt to select only patient-authored content during the data annotation process (see **Methods**). In our manual validation, the overall accuracy of 0.9394 accounted for any mislabeling of third-party narratives as patient-authored content. This type of mislabeling represents only a portion of the approximately 6% total mislabeling rate. Therefore, the influence of such errors on our findings is likely minimal, and their impact is already reflected in our sensitivity analysis and confidence interval estimates.

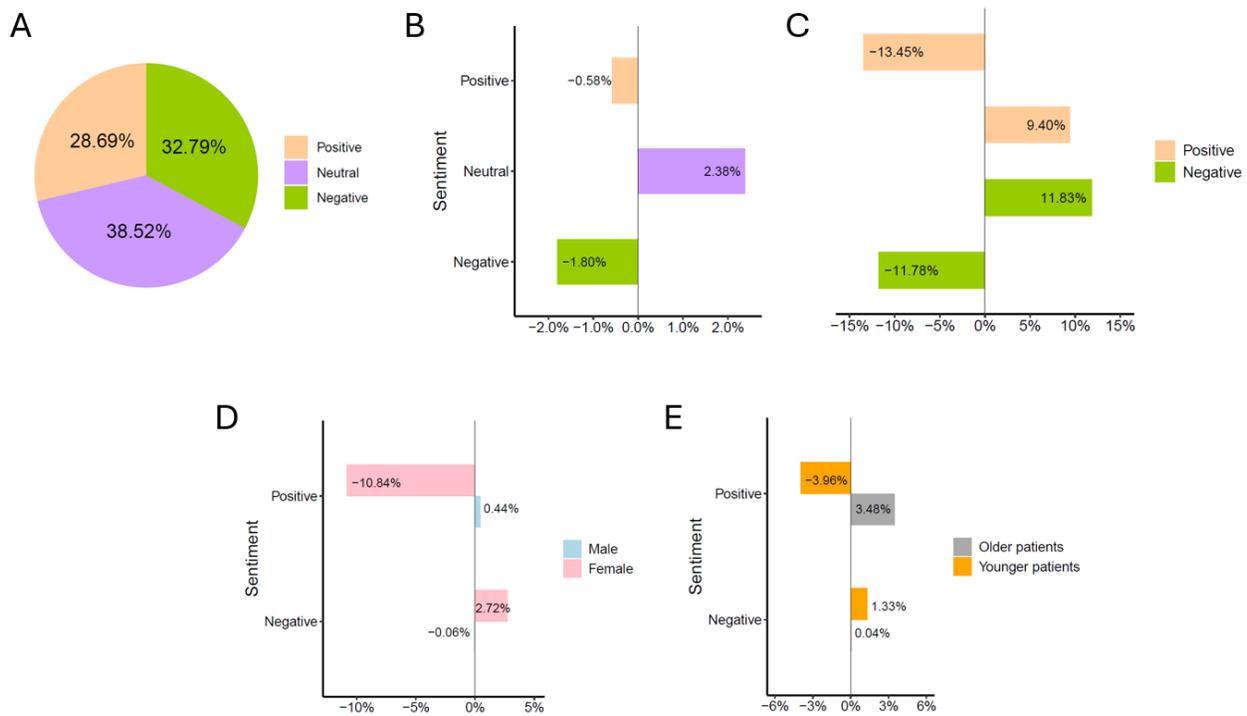

**Figure 5. Sentiment analysis of sarcoidosis patients. A.** Overall sentiment of all threads posted by sarcoidosis patients. **B.** Sentiment changes among patients after sarcoidosis diagnosis. **C.** Changes of positive and negative threads, averaged across patients. **D**. Changes of positive and negative threads after sarcoidosis diagnosis, separated by gender. **E**. Changes of positive and negative threads after sarcoidosis diagnosis, separated by age.

**Potential demographic bias.** Given that older adults are less likely to use Reddit, this age group may be underrepresented in our data. To assess the impact of this potential bias on our results, we conducted a weighting analysis using external references on the age distribution of sarcoidosis patients. First, we used the population-based study by Ungprasert et al.[5] as a real-world baseline for the age distribution among sarcoidosis patients. We then compared this baseline with the age distribution in our Reddit sample (**Supplementary Figure S2 and Supplementary Tables S15**). The comparison revealed that our Reddit sample skews younger compared to the population-based study. Next, we calculated weights for each age group using the following formula:

$$Weight = \frac{Percentage_{baseline}}{Percentage_{Reddit}}$$

We then re-analyzed the symptom and treatment data with and without weighting to evaluate the impact of the age differences. In the symptom weighting analysis, we focused on patients with both age and symptom data (n=125), multiplying the count of each symptom by the corresponding age group's weight. Similarly, in the treatment weighting analysis, we focused on patients with both age and treatment data (n=107), applying the appropriate weights.

**Supplementary Tables S16 and S17** present a comparison of the counts, percentages, and ranks with and without weighting for symptoms and treatments. For the 29 identified symptoms, the average count difference is 0.99, and the average rank difference is 1.38. For the 29 identified treatments, the average count difference is 0.77, and the average rank difference is 2.41. Importantly, the top 8 symptoms and top 9 treatments remained consistent regardless of weighting. This analysis indicates that while there are differences in age distributions between the Reddit sample and the population-based study, these differences do not significantly impact our primary findings related to symptoms and treatments. We anticipate similar stability in other analyses within this study, which we will explore further in future research. We also emphasize that potential sample bias, whether due to age or other factors, should be carefully examined in similar studies.

**Considerations for treatment strategies and clinical practice.** As previously demonstrated, phenotypes 1 and 3 are associated with better prognosis and shorter disease duration compared to phenotype 2 (**Figure 4D and 4E**). The treatment strategies used across these phenotypes suggest certain approaches that may be linked to improved outcomes. **Supplementary Table S18** summarizes the treatments used in each phenotype. We found that prednisone, methotrexate, and infliximab are commonly used across all three phenotypes. Notably, physical therapy is uniquely utilized in phenotypes 1 and 3—both of which are associated with better prognosis and shorter disease duration—indicating its potential efficacy in managing sarcoidosis. Patients in phenotype 2, who exhibited worse prognosis and longer disease duration, received a broader range of treatments not commonly used in the other groups, including leflunomide, methylprednisolone, mycophenolate, nonsteroidal anti-inflammatory drugs, and pulmonary rehabilitation. These additional treatments did not appear to correlate with better outcomes. We also analyzed the number of treatments used by each patient across the different phenotypes. As shown in **Supplementary Table S19**, patients in phenotype 2 received more types of treatments on average compared to those in phenotypes 1 and 3 (2.7 treatments vs. 1.9 treatments in both phenotypes 1 and 3). Additionally, the maximum number of treatments used by a single patient was higher in phenotype 2 (6 treatments) compared to phenotypes 1 and 3 (3 and 4 treatments, respectively). Given the worse prognosis and longer disease duration in phenotype 2, this suggests that multi-drug combination therapy may not be effective in managing the disease.

These findings should be interpreted with caution when applied to clinical practice. First, the observed relationships between prognosis, disease duration, and treatment strategies are correlational, not necessarily causal. The poorer prognosis in phenotype 2 may be due to the more severe baseline condition in this group rather than the treatments used. A randomized controlled trial would be necessary to establish causality. Second, the identification of phenotypes required complete data across all variables, leading to a relatively small sample size due to missing values. Consequently, these findings may not generalize well to the broader population due to statistical variability. Third, any clinical recommendations based on these results require external validation in an independent cohort to ensure their applicability. Future studies may address these limitations to develop more effective interventions.

**Sample size and generalizability.** While our dataset includes nearly seven hundred patients, which is relatively large for a rare disease, we acknowledge that it represents only a small fraction of the total sarcoidosis patient population. Therefore, caution should be taken when generalizing our findings to the broader patient population. A key objective of our study is to propose a novel framework and pipeline for understanding sarcoidosis and other rare diseases using social media data. The current moderate-sized sample is limited by the available content on Reddit. As the volume of content on Reddit continues to grow, we anticipate that future studies will be able to collect more data, resulting in a more representative sample of the sarcoidosis population. To achieve this, our analysis framework is designed to be adaptable to other rare diseases or larger online data sources that may contain significantly larger patient cohorts.

## Conclusion

Our study provides a comprehensive understanding of sarcoidosis through the analysis of social media data using Large Language Models (LLMs). Leveraging data from Reddit's sarcoidosis forum, we successfully extracted and analyzed information related to diagnosis, symptoms, treatments, prognosis, side effects, demographics, patient phenotype, and sentiment. Our findings confirm significant disparities in disease outcomes based on age and gender. The data-driven insights from social media have also allowed us to identify three distinct patient phenotypes, providing a deeper understanding of sarcoidosis that could lead to more tailored treatment approaches. The sentiment analysis of patient posts has revealed a moderate negative impact on mental health following diagnosis, particularly among females and younger individuals, underscoring the importance of psychological support in disease management strategies. Using cutting-edge artificial intelligence techniques, our findings have direct implications for improving personalized treatment and enhancing quality of care for individuals living with sarcoidosis. Furthermore, our study demonstrates the potential of social media data and advanced language models in health research, opening up promising avenues for future investigations into rare diseases and patient experiences.

There are several directions for future research that we plan to explore. First, future studies could incorporate data from additional social media platforms and online forums to gather a broader spectrum of patient experiences. This could enhance the diversity of the data and provide a more comprehensive understanding of sarcoidosis across different demographics and geographical locations. Second, conducting longitudinal studies to track changes in patient conditions over time could offer valuable insights into the progression of sarcoidosis and the long-term efficacy of treatments. This could also help understand how patient sentiments and community support evolve as the disease progresses. Third, integrating social media data with clinical trial data could provide real-world evidence that complements controlled study results, aiding in the development of new treatments and the improvement of existing ones. Lastly, based on the insights gained from social media analysis, there is potential to develop dedicated tools and platforms to better support sarcoidosis patients. These tools could provide tailored health information and guidance to healthcare professionals.

## Funding

This work was supported by National Science Foundation DMS-2413741, National Institute of Science HL134828, and Central Indiana Corporate Partnership AnalytiXIN Initiative.

## Data Availability

The data collected and analyzed in this study is available at Zenodo respiratory at https://doi.org/10.5281/zenodo.13370310.

# Appendix

## Supplementary Tables

**Supplementary Table S1.** Common sarcoidosis symptoms documented in medical literature.

| General symptoms | Pulmonary symptoms | Ocular symptoms | Dermatological symptoms | Cardiac symptoms | Neurological symptoms |
|---|---|---|---|---|---|
| Fever | Cough | Blurred vision or loss of vision | Growths under the skin around scars | Heart failure | Increased thirst or amounts of pee |
| Fatigue | Shortness of breath | Eye pain | Light or dark patches of skin | Fluttering heartbeat | Weak or paralyzed facial muscles |
| Joint pain | Chest pain | Red or swollen eyes | Raised reddish-purple sores | Irregular heartbeat | Headaches |
| Muscle aches or weakness | Wheezing | Sensitivity to light | Red and tender bumps on shins | | Seizures |
| Night sweats | | | | | |
| Swollen lymph nodes | | | | | |
| Unexplained weight loss | | | | | |
| Kidney stones | | | | | |

**Supplementary Table S2.** Common sarcoidosis medications and side effects recorded in clinical practice.

| Medication | Side effect |
|---|---|
| Colchicine, Minocycline, Tetracycline, Doxycycline, Pentoxifylline, Chloroquine, Rituximab, Infliximab, Golimumab, Adalimumab, Leflunomide, Azathioprine, Corticotropin, Prednisone, Cortisone, Methotrexate, Infliximab, Nonsteroidal, Anti-inflammatory drugs, Azathioprine, Hydroxychloroquine, Physical therapy, Defibrillator, Pulmonary rehabilitation, Antimalarials | Weight gain, Insomnia, Acne, Diabetes, High blood pressure, Glaucoma, Cataracts, Osteoporosis, Depression, Emotional irritability, Skin bruising, Hoarse voice, Thrush, Nausea, Dizziness, Gastrointestinal tract problems, Diarrhea |

**Supplementary Table S3.** Textual examples from Reddit text, along with corresponding model annotations for disease prognosis.

| Text | Prognosis |
|---|---|
| …I am really starting to feel better since I've took out stress and started to listen to my body. I go to bed when I need to, meditate daily, started to get my muscle mass back and work actively on restoring my vagal nerve and feel safe and relaxed in my body… | Good |
| …So my sarc is finally under control and my granulomas have started to shrink which is amazing news! It's been about 6 months maybe since I've had any symptoms at all… | Good |
| I have pulmonary sarcoidosis. I was told it was resolving on its own, and I was feeling better. I recently started having a flare up and I'm feeling so bad. Horrible body pain, nausea, light headed, chest pain, and headaches. Absolutely no appetite… | Moderate |
| I've had sarc for probably two or three years and my symptoms have been pretty mild. It did get into my lungs and I lost 10 - 15% of my lung capacity forever I guess. Other than that it's just the constant minor symptoms: aching joints, feeling feverish, small rashes etc… | Moderate |
| …I'm suffering mentally, physically, financially. I feel like less of a man as each day passes, having to ask my wife to do jobs that should be mine… | Poor |
| I demanded to get off of prednisone / methotrexate last summer and went to azathioprine with methotrexate. My ct scans are still showing increased pulmonary fibrosis 3 years and counting. Been on 4L of oxygen resting 6L in motion for 2.5yrs. Prednisone ballooned me from 200 to 280 in 1.2 years… | Poor |

**Supplementary Table S4.** Textual examples posted by sarcoidosis patients, along with the model annotated sentiment. Contents are general and not necessarily sarcoidosis-related.

| Text | Sentiment |
|---|---|
| I'm spending about 90 minutes a day with stretching, massage, and strength training. I feel like I'm winning the war, and am curious what a maintenance routine looks like to maintain good health once you are healed. | Positive |
| Hi everyone! I have been on keto for 3 months and I love it. I've lost 43lbs and have another 19 to go! My husband and I are starting to try to get pregnant next month, so my big concern is how to maintain my newly slim figure during/after the pregnancy. I just bought the book Real Food for Pregnancy and will read it right away… | Positive |
| Hey, has anyone of you Experience with Rituximab and (Small Fiber) Neuropathy / Painful Tingling? I was diagnosed with it when I was 30 and I'm going to start it soon and I'd be interested to know if anyone has experience with it to relieve this pain? | Moderate |
| I have two very tender spots near the collarbone on each side at what I think is my anterior scalene. I'm working them as trigger points, and just wanted to know if it's possible that's the brachial plexus. | Moderate |
| I don't understand why it's so hard people to understand that you need to actually try hard to get into grad programs. I'm going to cry if I get a B, because to get into the MD schools In my state, I need around a 3.7 at least. It's just infuriating when me and friends are talking about grades, and they refuse to realize, if I don't try my hardest, my degree will be pretty useless… | Negative |
| Is this normal? This is in the US. I'm about to have a heart attack over here, especially because I'm 99% sure it failed. | Negative |

**Supplementary Table S5. Prognosis by patient groups based on the number of symptoms.** n=269, p-value = 0.0188, $\chi^2$ test of independence.

|  |  | Prognosis | | |
| --- | --- | --- | --- | --- |
|  |  | Good | Moderate | Poor |
| Number of symptoms | 1 | 48 | 11 | 4 |
|  | 2 | 40 | 22 | 8 |
|  | 3 | 33 | 19 | 7 |
|  | >3 | 34 | 29 | 14 |

**Supplementary Table S6. Prognosis by patient groups based on the number of symptom categories.** n=269, p-value = 0.0006, $\chi^2$ test of independence.

|  |  | Prognosis | | |
| --- | --- | --- | --- | --- |
|  |  | Good | Moderate | Poor |
| Number of symptom categories | 1 | 73 | 29 | 6 |
|  | 2 | 64 | 28 | 16 |
|  | >2 | 18 | 24 | 11 |

**Supplementary Table S7. Prognosis by patient groups with a single symptom category.** n=108, p-value = 0.3376, $\chi^2$ test of independence.

|  |  | Prognosis | | |
| --- | --- | --- | --- | --- |
|  |  | Good | Moderate | Poor |
| Single symptom category | General | 41 | 16 | 1 |
|  | Pulmonary | 14 | 5 | 1 |
|  | Ocular | 6 | 1 | 1 |
|  | Dermatological | 4 | 1 | 1 |
|  | Cardiac | 4 | 5 | 2 |
|  | Neurological | 4 | 1 | 0 |

**Supplementary Table S8. Symptoms reported by sarcoidosis patients (n=485).** Some patients reported multiple symptoms.

| symptom | Percentage (%) | Count |
|---|---|---|
| Fatigue | 35.26 | 171 |
| Swollen lymph nodes | 31.34 | 152 |
| Shortness of breath | 24.33 | 118 |
| Joint pain | 22.27 | 108 |
| Cough | 20.00 | 97 |
| Chest pain | 17.94 | 87 |
| Blurred vision or loss of vision | 14.85 | 72 |
| Headaches | 11.96 | 58 |
| Unexplained weight loss | 10.52 | 51 |
| Muscle aches or weakness | 9.48 | 46 |
| Fluttering heartbeat | 8.04 | 39 |
| Growths under the skin around scars or tattoos | 7.84 | 38 |
| Irregular heartbeat | 7.22 | 35 |
| Eye pain | 7.01 | 34 |
| Sensitivity to light | 7.01 | 34 |
| Night sweats | 6.60 | 32 |
| Red or swollen eyes | 6.39 | 31 |
| Increased thirst or amounts of pee | 4.95 | 24 |
| Kidney stones | 4.95 | 24 |
| Light or dark patches of skin | 4.95 | 24 |
| Wheezing | 4.74 | 23 |
| Raised reddish-purple sores or rash across nose or cheeks | 4.33 | 21 |
| Fever | 3.92 | 19 |
| Heart failure | 3.92 | 19 |
| Red tender bumps on shins | 3.92 | 19 |
| Seizures | 2.68 | 13 |
| Weak or paralyzed facial muscles | 2.68 | 13 |

**Supplementary Table S9. Prognosis of all patients (n=321) and those who used the top 3 medications.**

|  | Prognosis | Percentage (%) | Count |
|---|---|---|---|
| Overall | Good | 61.37 | 197 |
|  | Moderate | 26.17 | 84 |
|  | Poor | 12.46 | 40 |
| Prednisone | Good | 56.80 | 96 |
|  | Moderate | 33.14 | 56 |
|  | Poor | 10.06 | 17 |
| Methotrexate | Good | 37.93 | 33 |
|  | Moderate | 49.43 | 43 |
|  | Poor | 12.64 | 11 |
| Infliximab | Good | 45.45 | 15 |
|  | Moderate | 48.48 | 16 |
|  | Poor | 6.06 | 2 |

**Supplementary Table S10. Prognosis distributions of patient age and gender groups.**

|  |  | Prognosis | Percentage (%) | Count |
|---|---|---|---|---|
| Age | Young | Good | 58.54 | 24 |
|  |  | Moderate | 21.95 | 9 |
|  |  | Poor | 19.51 | 8 |
|  | Old | Good | 42.31 | 22 |
|  |  | Moderate | 44.23 | 23 |
|  |  | Poor | 13.46 | 7 |
| Gender | Male | Good | 48.57 | 17 |
|  |  | Moderate | 34.29 | 12 |
|  |  | Poor | 17.14 | 16 |
|  | Female | Good | 62.86 | 22 |
|  |  | Moderate | 25.71 | 9 |
|  |  | Poor | 11.43 | 4 |

**Supplementary Table S11. Prognosis distributions of three patient phenotypes.**

|  | Prognosis | Percentage (%) | Count |
|---|---|---|---|
| Phenotype 1 | Good | 58.33 | 7 |
|  | Moderate | 16.67 | 2 |
|  | Poor | 25.00 | 3 |
| Phenotype 2 | Good | 31.25 | 5 |
|  | Moderate | 56.25 | 9 |
|  | Poor | 12.50 | 2 |
| Phenotype 3 | Good | 61.11 | 11 |
|  | Moderate | 33.33 | 6 |
|  | Poor | 5.56 | 1 |

**Supplementary Table S12. Gender distributions of three patient phenotypes.**

|  | Gender | Percentage (%) | Count |
|---|---|---|---|
| Phenotype 1 | Male | 63.64 | 14 |
|  | Female | 36.36 | 8 |
| Phenotype 2 | Male | 50.00 | 11 |
|  | Female | 50.00 | 11 |
| Phenotype 3 | Male | 29.63 | 8 |
|  | Female | 70.37 | 19 |

**Supplementary Table S13. Sentiment of threads posted by sarcoidosis patients, categorized by overall, before, and after diagnosis.** The threads posted exactly at diagnosis time were not counted.

|  | Sentiment | Percentage (%) | Count |
|---|---|---|---|
| Overall | Positive | 28.69 | 6914 |
|  | Neutral | 38.52 | 9285 |
|  | Negative | 32.79 | 7904 |
| Before diagnosis | Positive | 29.09 | 4141 |
|  | Neutral | 37.76 | 5374 |
|  | Negative | 33.15 | 4178 |
| After diagnosis | Positive | 28.51 | 2724 |
|  | Neutral | 40.13 | 3834 |
|  | Negative | 31.35 | 2995 |

**Supplementary Table S14. 95% confidence intervals for percentage and count of top three treatments and side effects.**

| Treatment | 95% CI (percentage) | 95% CI (Count) | Side effect | 95% CI (percentage) | 95% CI (Count) |
|---|---|---|---|---|---|
| Prednisone | (69.86%, 74.79%) | (255,273) | Weight gain | (45.27%, 50.75%) | (91, 102) |
| Methotrexate | (38.08%, 43.29%) | (139, 158) | Insomnia | (36.82%, 42.29%) | (74, 85) |
| Infliximab | (18.63%, 24.38%) | (68, 89) | Diabetes | (34.33%, 39.80%) | (69, 80) |

**Supplementary Table S15. Age distributions of sarcoidosis patients reported in a population-based study by Ungprasert et al. and among patient texts on Reddit used in this study.**

| | Ungprasert et al. (n=448) | | Reddit (n=144) | | |
|---|---|---|---|---|---|
| Age group | Count | Percentage (%) | Count | Percentage (%) | Weight |
| <29 | 72 | 16.1 | 33 | 22.9 | 0.701 |
| 30-39 | 116 | 25.9 | 52 | 36.1 | 0.717 |
| 40-49 | 114 | 25.4 | 30 | 20.8 | 1.221 |
| 50-59 | 89 | 19.9 | 20 | 13.9 | 1.43 |
| 60-69 | 36 | 8.0 | 7 | 4.9 | 1.653 |
| >70 | 21 | 4.7 | 2 | 1.4 | 3.375 |

**Supplementary Table S16. Symptom counts, percentages, and ranks before and after age weighting.**
The sample size (n=125) includes patients with both age and symptom information.

| Symptom | Original | | | Weighted | | | Difference | |
|---|---|---|---|---|---|---|---|---|
| | Count | Percentage (%) | Rank | Count | Percentage (%) | Rank | Count | Rank |
| Fatigue | 47 | 37.6 | 1 | 51 | 40.8 | 1 | 3.2 | 0 |
| Shortness of breath | 47 | 37.6 | 2 | 51 | 40.8 | 2 | 3.2 | 0 |
| Swollen lymph nodes | 36 | 28.8 | 3 | 33 | 26.4 | 3 | 2.4 | 0 |
| Chest pain | 31 | 24.8 | 4 | 29 | 23.2 | 5 | 1.6 | 1 |
| Joint pain | 30 | 24 | 5 | 30 | 24 | 4 | 0 | 1 |
| Cough | 27 | 21.6 | 6 | 25 | 20 | 6 | 1.6 | 0 |
| Blurred vision or loss of vision | 20 | 16 | 7 | 21 | 16.8 | 8 | 0.8 | 1 |
| Headaches | 19 | 15.2 | 8 | 23 | 18.4 | 7 | 3.2 | 1 |
| Unexplained weight loss | 14 | 11.2 | 9 | 13 | 10.4 | 11 | 0.8 | 2 |
| Fluttering heartbeat | 13 | 10.4 | 10 | 13 | 10.4 | 10 | 0 | 0 |
| Night sweats | 13 | 10.4 | 11 | 11 | 8.8 | 15 | 1.6 | 4 |
| Eye pain | 12 | 9.6 | 12 | 13 | 10.4 | 9 | 0.8 | 3 |
| Muscle aches or weakness | 12 | 9.6 | 13 | 11 | 8.8 | 14 | 0.8 | 1 |
| Increased thirst or amounts of pee | 11 | 8.8 | 14 | 10 | 8 | 17 | 0.8 | 3 |
| Irregular heartbeat | 11 | 8.8 | 15 | 11 | 8.8 | 13 | 0 | 2 |
| Heart failure | 10 | 8 | 16 | 11 | 8.8 | 12 | 0.8 | 4 |
| Wheezing | 10 | 8 | 17 | 10 | 8 | 18 | 0 | 1 |
| Sensitivity to light | 9 | 7.2 | 18 | 11 | 8.8 | 16 | 1.6 | 2 |
| Red or swollen eyes | 7 | 5.6 | 19 | 9 | 7.2 | 19 | 1.6 | 0 |
| Growths under the skin around scars or tattoos | 6 | 4.8 | 20 | 5 | 4 | 22 | 0.8 | 2 |
| Kidney stones | 6 | 4.8 | 21 | 6 | 4.8 | 20 | 0 | 1 |
| Weak or paralyzed facial muscles | 6 | 4.8 | 22 | 6 | 4.8 | 21 | 0 | 1 |
| Fever | 4 | 3.2 | 23 | 3 | 2.4 | 24 | 0.8 | 1 |
| Red tender bumps on shins | 4 | 3.2 | 24 | 3 | 2.4 | 26 | 0.8 | 2 |
| Seizures | 4 | 3.2 | 25 | 3 | 2.4 | 27 | 0.8 | 2 |
| Light or dark patches of skin | 3 | 2.4 | 26 | 4 | 3.2 | 23 | 0.8 | 3 |
| Raised reddish-purple sores or rash across nose or cheeks | 3 | 2.4 | 27 | 3 | 2.4 | 25 | 0 | 2 |
| Painful Tingling | 1 | 0.8 | 28 | 1 | 0.8 | 28 | 0 | 0 |
| Tremors | 1 | 0.8 | 29 | 1 | 0.8 | 29 | 0 | 0 |
| **Average difference** | | | | | | | 0.99 | 1.38 |

**Supplementary Table S17. Treatment counts, percentages, and ranks before and after age weighting.**
The sample size (n=107) includes patients with both age and treatment information.

|  | Original | | | Weighted | | | Difference | |
| --- | --- | --- | --- | --- | --- | --- | --- | --- |
| Symptom | Count | Percentage (%) | Rank | Count | Percentage (%) | Rank | Count | Rank |
| Prednisone | 84 | 78.5 | 1 | 89 | 83.2 | 1 | 4.7 | 0 |
| Methotrexate | 48 | 44.9 | 2 | 52 | 48.6 | 2 | 3.7 | 0 |
| Infliximab | 21 | 19.6 | 3 | 22 | 20.6 | 3 | 0.9 | 0 |
| Azathioprine | 10 | 9.3 | 4 | 10 | 9.3 | 5 | 0 | 1 |
| Defibrillator | 10 | 9.3 | 5 | 11 | 10.3 | 4 | 0.9 | 1 |
| Prednisolone | 8 | 7.5 | 6 | 8 | 7.5 | 7 | 0 | 1 |
| Physical Therapy | 6 | 5.6 | 7 | 6 | 5.6 | 9 | 0 | 2 |
| Hydroxychloroquine | 5 | 4.7 | 8 | 8 | 7.5 | 6 | 2.8 | 2 |
| Leflunomide | 3 | 2.8 | 9 | 6 | 5.6 | 8 | 2.8 | 1 |
| Corticotropin | 2 | 1.9 | 10 | 2 | 1.9 | 12 | 0 | 2 |
| Cortisone | 2 | 1.9 | 11 | 1 | 0.9 | 19 | 0.9 | 8 |
| Humira | 2 | 1.9 | 12 | 2 | 1.9 | 13 | 0 | 1 |
| Methylprednisolone | 2 | 1.9 | 13 | 1 | 0.9 | 21 | 0.9 | 8 |
| Nonsteroidal Anti-Inflammatory Drugs | 2 | 1.9 | 14 | 4 | 3.7 | 10 | 1.9 | 4 |
| Rituximab | 2 | 1.9 | 15 | 1 | 0.9 | 27 | 0.9 | 12 |
| Steroids | 2 | 1.9 | 16 | 2 | 1.9 | 15 | 0 | 1 |
| Adalimumab | 1 | 0.9 | 17 | 1 | 0.9 | 16 | 0 | 1 |
| Advair | 1 | 0.9 | 18 | 1 | 0.9 | 17 | 0 | 1 |
| Albuterol | 1 | 0.9 | 19 | 2 | 1.9 | 11 | 0.9 | 8 |
| Cellcept | 1 | 0.9 | 20 | 1 | 0.9 | 18 | 0 | 2 |
| Hydroxyzine | 1 | 0.9 | 22 | 1 | 0.9 | 20 | 0 | 2 |
| Minocycline | 1 | 0.9 | 23 | 1 | 0.9 | 22 | 0 | 1 |
| Mycophenolate | 1 | 0.9 | 24 | 1 | 0.9 | 23 | 0 | 1 |
| Naproxen | 1 | 0.9 | 25 | 1 | 0.9 | 24 | 0 | 1 |
| Dexamethasone | 1 | 0.9 | 26 | 1 | 0.9 | 25 | 0 | 1 |
| Pulmonary Rehabilitation | 1 | 0.9 | 27 | 1 | 0.9 | 26 | 0 | 1 |
| Tnfa Inhibitor | 1 | 0.9 | 28 | 1 | 0.9 | 28 | 0 | 0 |
| Zessly | 1 | 0.9 | 29 | 1 | 0.9 | 29 | 0 | 0 |
| **Average difference** | | | | | | | 0.77 | 2.41 |

**Supplementary Table S18. Treatments used by patients in the three phenotypes.**

| Phenotype 1 (n=22) | | | Phenotype 2 (n=22) | | | Phenotype 3 (n=27) | | |
|---|---|---|---|---|---|---|---|---|
| Treatment | Count | Percentage (%) | Treatment | Count | Percentage (%) | Treatment | Count | Percentage (%) |
| Prednisone | 10 | 58.8 | Prednisone | 16 | 84.2 | Prednisone | 15 | 78.9 |
| Methotrexate | 8 | 47.1 | Methotrexate | 10 | 52.6 | Methotrexate | 10 | 52.6 |
| Defibrillator | 3 | 17.6 | Infliximab | 7 | 31.8 | Infliximab | 4 | 21.1 |
| Infliximab | 3 | 17.6 | Azathioprine | 5 | 26.3 | Azathioprine | 2 | 10.5 |
| Physical therapy | 2 | 11.8 | Prednisolone | 3 | 15.8 | Hydroxychloroquine | 2 | 10.5 |
| Prednisolone | 2 | 11.8 | Hydroxychloroquine | 2 | 10.5 | Physical therapy | 2 | 10.5 |
| Adalimumab | 1 | 5.9 | Defibrillator | 2 | 10.5 | Naproxen | 1 | 5.3 |
| Minocycline | 1 | 5.9 | Leflunomide | 2 | 10.5 | | | |
| Rituximab | 1 | 5.9 | Methylprednisolone | 1 | 5.3 | | | |
| Steroids | 1 | 5.9 | Mycophenolate | 1 | 5.3 | | | |
| | | | Nonsteroidal anti-inflammatory drugs | 1 | 5.3 | | | |
| | | | Pulmonary rehabilitation | 1 | 5.3 | | | |

**Supplementary Table S19. Summary statistics of treatment numbers per patient in the three phenotypes.**

| | | Phenotype 1 | Phenotype 2 | Phenotype 3 |
|---|---|---|---|---|
| | Minimum | 1 | 1.9 | 3 |
| Treatment number per patient | Mean | 1 | 2.7 | 6 |
| | Maximum | 1 | 1.9 | 4 |

# Supplementary Figures

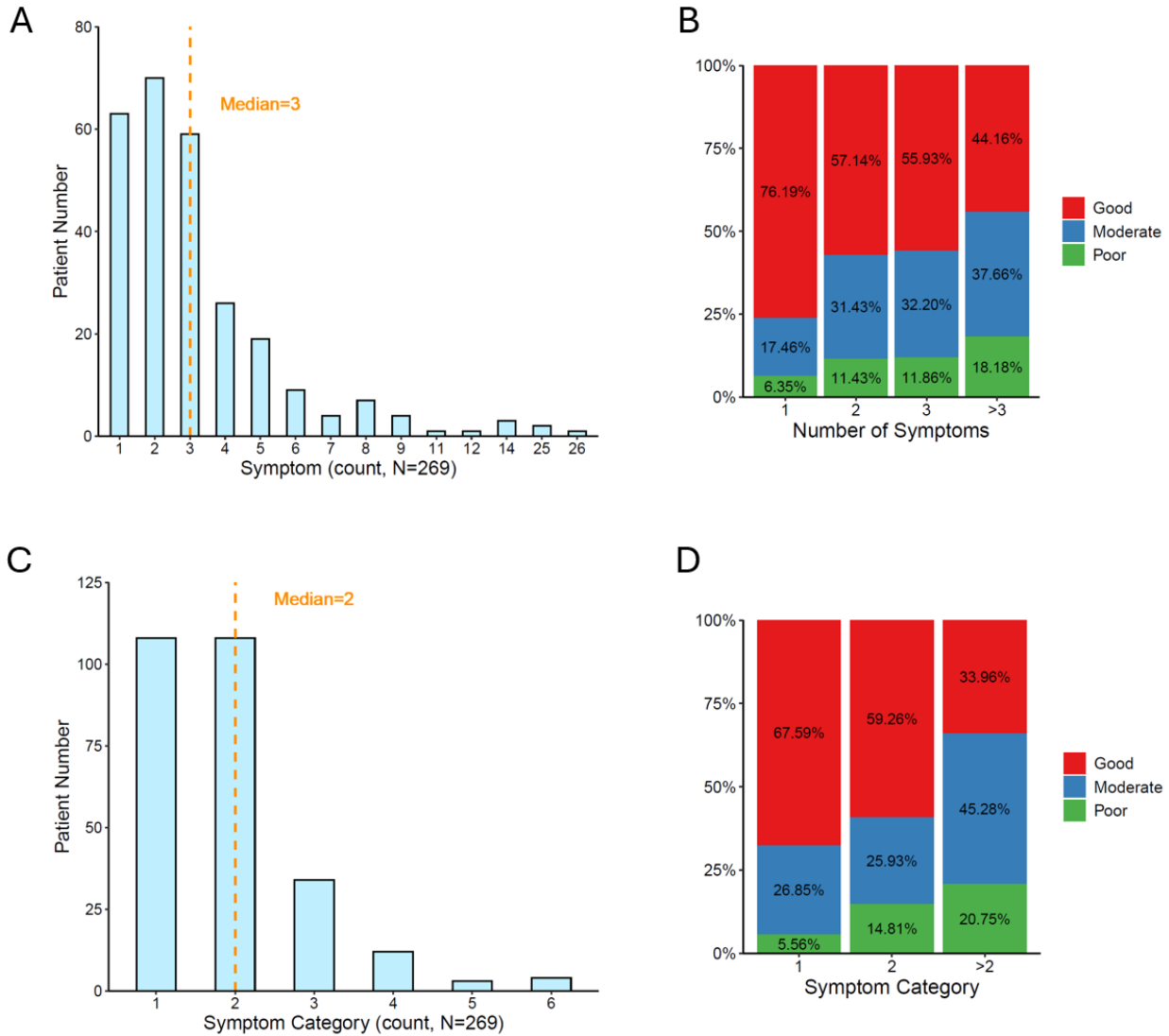

**Supplementary Figure S1. Relationship between symptoms and prognosis. A.** Distribution of the number of symptoms among sarcoidosis patients. **B.** Prognosis distribution of patient groups based on the number of symptoms. p-value = 0.0188, $\chi^2$ test of independence. **C.** Distribution of the number of symptom categories among sarcoidosis patients. **D.** Prognosis distribution of patient groups based on the number of symptom categories. p-value = 0.0006, $\chi^2$ test of independence.

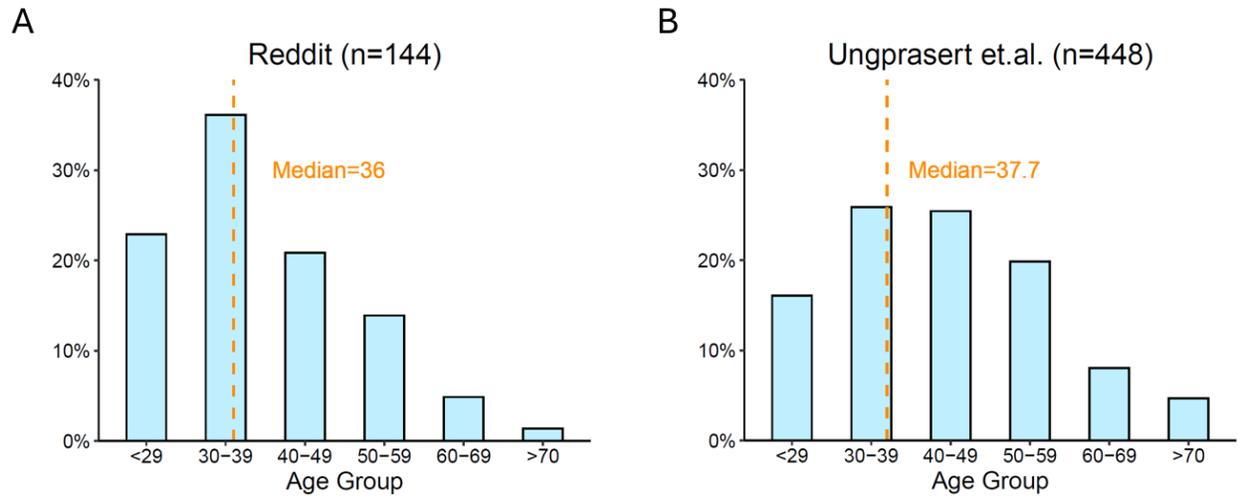

**Supplementary Figure S2. Age distributions of sarcoidosis patients. A.** Among patient texts on Reddit used in this study. **B.** As reported in a population-based study by Ungprasert et al.